\documentclass[10pt,twocolumn,letterpaper]{article}

\usepackage{cvpr}
\usepackage{times}
\usepackage{epsfig}
\usepackage{graphicx}

\usepackage{amsmath,amsfonts,bm}

\def\eqref#1{equation~\ref{#1}}

\def\1{\bm{1}}

\DeclareMathAlphabet{\mathsfit}{\encodingdefault}{\sfdefault}{m}{sl}
\SetMathAlphabet{\mathsfit}{bold}{\encodingdefault}{\sfdefault}{bx}{n}

\DeclareMathOperator*{\argmin}{arg\,min}

\usepackage{algorithm}
\usepackage{algorithmic}
\usepackage{url}
\usepackage{subcaption}
\usepackage{amsmath, amsthm, amssymb}
\usepackage[ansinew]{inputenc}
\usepackage{multirow}
\usepackage{booktabs}
\usepackage[flushleft]{threeparttable}
\usepackage{tabulary}
\usepackage{wrapfig}
\usepackage{dblfloatfix}
\usepackage[pagebackref=true,breaklinks=true,letterpaper=true,colorlinks,bookmarks=false]{hyperref}

\newcommand{\mmL}{\mathcal{L}}

\newcommand{\mldas}{MiLeNAS}

\cvprfinalcopy 

\def\cvprPaperID{6901} 

\begin{document}
\title{MiLeNAS: Efficient Neural Architecture Search via Mixed-Level Reformulation}
\author{Chaoyang He\textsuperscript{1}\thanks{Equal contribution}\quad Haishan Ye\textsuperscript{2}\footnotemark[1]\quad Li Shen\textsuperscript{3}\quad Tong Zhang\textsuperscript{4}\\ 
\textsuperscript{1}University of Southern California\quad \textsuperscript{2}The Chinese University of Hong Kong, Shenzhen \\ \quad\textsuperscript{3}Tencent AI Lab\quad \textsuperscript{4}Hong Kong University of Science and Technology\\
\tt\small chaoyang.he@usc.edu\quad hsye\_cs@outlook.com\quad lshen.lsh@gmail.com\quad  tongzhang@tongzhang-ml.org
}


\maketitle
\begin{abstract}
Many recently proposed methods for Neural Architecture Search (NAS) can be formulated as bilevel optimization. For efficient implementation, its solution requires approximations of second-order methods. In this paper, we demonstrate that gradient errors caused by such approximations lead to suboptimality, in the sense that the optimization procedure fails to converge to a (locally) optimal solution. To remedy this, this paper proposes \mldas, a mixed-level reformulation for NAS that can be optimized efficiently and reliably. It is shown that even when using a simple first-order method on the mixed-level formulation, \mldas\ can achieve a lower validation error for NAS problems. Consequently, architectures obtained by our method achieve consistently higher accuracies than those obtained from bilevel optimization. Moreover, \mldas\ proposes a framework beyond DARTS. It is upgraded via model size-based search and early stopping strategies to complete the search process in around 5 hours. Extensive experiments within the convolutional architecture search space validate the effectiveness of our approach.
\end{abstract}
\section{Introduction}
The success of deep learning in computer vision heavily depends on novel neural architectures \cite{he2016deep,huang2017densely}.
However, most widely-employed architectures are developed manually, making them time-consuming and error-prone. Thus, there has been an upsurge of research interest in neural architecture search (NAS), which automates the manual process of architecture design \cite{Bello2016Neural,real2017large}. There are three major methods for NAS: evolutionary algorithms \cite{real2017large,elsken2018efficient}, reinforcement learning-based methods \cite{Bello2016Neural,pham2018efficient}, and gradient-based methods \cite{liu2018darts,xie2018snas,elsken2018efficient,luo2018neural}. Developing optimization methods for the gradient-based NAS is promising since it achieves state-of-the-art performances on CNNs with less than one GPU day \cite{liu2018darts,dong2019searching}.

Formally, gradient-based methods can be formulated as a bilevel optimization problem \cite{liu2018darts}: 
\begin{align}
&\min_{\alpha}\quad \mmL_{\mathrm{val}} (w^*(\alpha), \alpha) \label{eq:bi_prob}\\
&\;\;\mathrm{s.t.}\quad w^*(\alpha) = \argmin_w \;\mmL_{\mathrm{tr}}(w, \alpha) \label{eq:bi_constraint}
\end{align}
where $w$ represents the network weight and $\alpha$ determines the neural architecture.  $\mmL_{tr} (w, \alpha)$ and $\mmL_{val} (w, \alpha)$ denote the losses with respect to training data and validation data with $w$ and $\alpha$, respectively.
Though bilevel optimization can accurately describe the NAS problem, 
it is difficult to solve, as obtaining $w^*(\alpha)$ in Equation \ref{eq:bi_constraint} requires one to completely train a network for each update of $\alpha$.
Current methods used in NAS to solve bilevel optimization are heuristic, and $w^*(\alpha)$ in Equation \ref{eq:bi_constraint} is not satisfied due to first-order or second-order approximation  \cite{liu2018darts,dong2019searching}. The second-order approximation has a superposition effect in that it builds upon the one-step approximation of $w$, causing gradient error and deviation from the true gradient.

Single-level optimization is another method used to solve the NAS problem and is defined as:
\begin{align}
\min_{w,\alpha} \mmL_{\mathrm{tr}}(w,\alpha),
\end{align}
which can be solved efficiently by stochastic gradient descent. 
However, single-level optimization commonly leads to overfitting with respect to $\alpha$, meaning that it cannot guarantee that the validation loss $\mmL_{\mathrm{val}}(w,\alpha)$ is sufficiently small.
This directly contradicts the objective of NAS, which is to minimize the validation loss to find the optimal structures.
Therefore, single-level optimization is insufficient for NAS. 

In this work, we propose mixed-level optimization, which incorporates both bilevel and single-level optimization schemes.
Rather than minimizing the validation loss with respect to $\alpha$ with the fully trained weights $w^*(\alpha)$ as in Equation \ref{eq:bi_constraint}, or directly minimizing $\alpha$ over the training loss, we minimize both the training loss and validation loss with respect to $\alpha$, and the training loss with respect to $w$, simultaneously. 
Note that when the hyperparameter $\lambda$ (Equation \ref{eq:alpha}) of our mixed-level optimization is set to zero, our mixed-level optimization method degrades to the single-level optimization. Alternatively, if $\lambda$ approaches infinity, our method becomes the bilevel optimization. Since we mix single-level and bilevel optimizations, we call our method \mldas, \emph{Mixed-Level optimization} based NAS.

\mldas\ can search with more stability and at faster speeds, and can find a better architecture with higher accuracy.
First, it has a computational efficiency similar to that of single-level optimization, but it is able to mitigate the overfitting issue.
Second, it can fully exploit both training data and validation data to update $\alpha$ and simultaneously avoid the gradient error caused by the approximation in the bilevel second-order methinod. Furthermore, \mldas\ upgrades the general DARTS framework \cite{liu2018darts}. In this framework, we demonstrate its versatility in two search space settings (DARTS  and GDAS \cite{dong2019searching}). Notably, this framework further introduces the model size-based search and early stopping strategies to largely accelerate the search speed (more details will be presented in Sections \ref{sec:beyond} and \ref{sec:strategy}). 

Extensive experiments validate the effectiveness of \mldas. 
We first correlate \mldas\ with single-level and bilevel methods by comparing their respective gaps between the training accuracy and the evaluation accuracy. 
The results show that \mldas\ can overcome overfitting, and that single-level and bilevel optimizations are special cases of \mldas.
Furthermore, \mldas\ achieves a better validation accuracy three times faster than bi-level optimization.
Evaluations on searched architectures show that \mldas\ reaches an error rate of $2.51\% \pm 0.11\%$ (best: $2.34\%$) on CIFAR-10, largely exceeding bilevel optimization methods (DARTS-$2.76\%$, GDAS-$2.82\%$). The transferability evaluation on ImageNet shows that \mldas\ has a top-1 error rate of 24.7\% and a top-5 error rate of 7.6\%, exceeding bilevel optimization methods by around 1\% to 2\%. 
Moreover, we demonstrate that \mldas\ is generic by applying it to the sampling-based search space. Finally, experiments with the model size-based and early stopping strategies introduced by the \mldas\ framework further provide several benefits in neural architecture design and accelerate the search speed to 5 hours.

We summarize our contributions as follows:
\begin{itemize}
    \item We propose a novel solution to the NAS problem by reformulating it as mixed-level optimization instead of bilevel optimization, alleviating the gradient error caused by approximation in bilevel optimization. This leads to a reliable first-order method as efficient as that of the single-level method.
    \item \mldas\ can search for better architectures with faster convergence rates. Extensive experiments on image classification demonstrate that \mldas\ can achieve a lower validation error at a search time three times shorter than that of bilevel optimization.
    \item \mldas\ introduces a NAS framework beyond DARTS. This framework demonstrates that \mldas\ is a generic framework for gradient-based NAS problems by demonstrating its versatility in sampling-based methods in obtaining better architectures. 
    \item The \mldas\ framework also introduces a model sized-based search strategy and an early stopping strategy to speed up the search process, and it also provides insights into neural architecture design.
\end{itemize}
We release the source code of MiLeNAS at \url{http://github.com/chaoyanghe/MiLeNAS}.
\section{Related Works}
While the deep architectures \cite{simonyan2014very,he2016deep,huang2017densely,hu2018squeeze} for convolutional neural networks (CNNs) are capable of tackling a wide range of visual tasks \cite{krizhevsky2012imagenet,toshev2014deeppose,long2015fully,ren2015faster},  neural architecture search (NAS) has attracted widespread attention due to its advantages over manually designed architectures.
There are three primary methods for NAS.
The first method relies on evolutionary algorithms \cite{real2017large,elsken2018efficient,yang2019cars}. These algorithms can simultaneously optimize
architectures and network weights.
However, their demand for enormous computational resources makes them highly restrictive (e.g., AmoebaNet \cite{real2019regularized} requires 3150 GPU days). The second method, reinforcement learning (RL) based NAS, formulates the design process of a neural network as a sequence of actions and regards the model accuracy as a reward \cite{Bello2016Neural,pham2018efficient}. 
The third method is gradient-based \cite{liu2018darts,xie2018snas,elsken2018efficient,luo2018neural,dong2019searching}, which relaxes the categorical design choices to continuous variables and then leverages the efficient gradient back-propagation so that it can finish searching within as little as several GPU days. Our work is related to this category, as we aim to further improve its efficiency and effectiveness.

Besides, several new NAS algorithms have been proposed to improve NAS from different perspectives. For example, task-agnostic NAS is proposed for the multi-task learning framework \cite{NAS_MTL};  releasing the constraints of hand-designed heuristics \cite{iccv2019hmnas} or alleviating the gap between the search accuracy and the evaluation accuracy are also promising directions \cite{chen2019progressive,li2019sgas}. Moreover, recent proposed NAS methods achieve a higher accuracy than our method \cite{nayman2019xnas,hundt2019sharpdarts,cai2018proxylessnas}. However, their improvements are due to novel searching spaces or searching strategies rather than a fundamental and generic optimization method. 
\section{Proposed Method}
MiLeNAS aims to search for better architectures efficiently. 
In this section, we first introduce mixed-level reformulation and propose MiLeNAS first-order and second-order methods for Neural Architecture Search. We then explain the benefits of MiLeNAS through theoretical analysis, which compares MiLeNAS with DARTS. Finally, we introduce the MiLeNAS framework and present additional benefits inspired by mixed-level optimization, including versatility in different search spaces, model size-based search, and early stopping strategy.

\subsection{Mixed-Level Reformulation}
\label{sec:moo}

We derive the mixed-level optimization from the single-level optimization, aiming to reduce $\alpha$ overfitting by considering both the training and validation losses. 
First, the single-level optimization problem is defined as:
\begin{equation}
\label{eq:method-train}
\min_{w,\alpha} \mmL_{tr} (w, \alpha)
:\equiv \min_\alpha \mmL_{tr}(w^*(\alpha),\alpha),
\end{equation}
where $\mmL_{tr} (w, \alpha)$ denotes the loss with respect to training data. When training neural network weights $w$, methods such as dropout are used to avoid overfitting with respect to $w$.
However, directly minimizing Equation \ref{eq:method-train} to obtain the optimal weight and architecture parameter may lead to overfitting with respect to $\alpha$. Because $\alpha$ solely depends on the training data, when it is optimized, there is a disparity between $\mmL_{tr} (w, \alpha)$ and $\mmL_{val} (w, \alpha)$.
Thus, the objective function defined in Equation \ref{eq:method-train} is inadequate for neural network search.

To alleviate the overfitting problem of $\alpha$, we resort to the most popular regularization method and use $\mmL_{val} (w, \alpha)$ as the regularization term.
Specifically, we minimize Equation \ref{eq:method-train} subject to the constraint
\begin{equation*}
    \mmL_{\mathrm{val}}(w^*(\alpha),\alpha) \leq \mmL_{\mathrm{tr}}(w^*(\alpha),\alpha) + \delta, 
\end{equation*}
where $\delta$ is a constant scalar.
The above constraint imposes that the validation loss could not be much larger than the training loss.
By the Lagrangian multiplier method, we minimize
\begin{gather*}
    w^*(\alpha)=\arg\min_w  \mmL_{tr} (w, \alpha), \\
    \min_{\alpha} (1-\lambda')\mmL_{tr} (w^*(\alpha), \alpha) + \lambda' \mmL_{val} (w^*(\alpha), \alpha) - \lambda'\delta, \\ 
   \quad 0\leq\lambda'\leq 1.
\end{gather*}
Because $\delta$ is a constant which does not affect the minimization, after normalizing the parameter before $\mmL_{tr} (w(\alpha), \alpha)$ to 1, we obtain the following mixed-level optimization using Equation \ref{eq:method-train}:
\begin{align}
&\min_{\alpha,w} 
\left[ \mmL_{tr}(w^*(\alpha),\alpha) + \lambda \mmL_{val}(w^*(\alpha),\alpha)\right] , \label{eq:alpha}
\end{align}
where $\lambda$ is a non-negative regularization parameter that balances the importance of the training loss and validation loss.
This is different from the bilevel optimization Equations \ref{eq:bi_prob} and \ref{eq:bi_constraint} and single-level optimization in Equation \ref{eq:method-train}.
Therefore, by taking the underlying relation between the training loss and validation loss into account, our mixed-level optimization can alleviate the overfitting issue and search for architectures with higher accuracy than single-level and bilevel optimizations.


We then apply the first-order method (stochastic gradient descent) to solve Equation \ref{eq:alpha} as follows:
\begin{equation}\label{eq:mixed_level}
    \begin{aligned}
&w = w - \eta_w \nabla_w \mmL_{\mathrm{tr}}(w, \alpha), \\
  &\alpha = \alpha - \eta_\alpha \left(\nabla_\alpha  \mmL_{\mathrm{tr}}(w, \alpha) +  \lambda \nabla_\alpha \mmL_{\mathrm{val}} (w, \alpha)\right),
    \end{aligned}
\end{equation}
where $\eta_w$ and $\eta_\alpha$ are step sizes related to $w$ and $\alpha$, respectively. 
Based on this \mldas\ first-order method, we can utilize the finite approximation to derive \mldas\ second-order method as follows: \\
1. $w=w-\eta_{w} \nabla_{w} L_{\operatorname{tr}}(w, \alpha), $ \\
2. Update $\alpha$ as follows:
\begin{equation}
\small
\begin{aligned}
&\alpha =\alpha-\eta_{\alpha}\\&\cdot\left[ \left( \nabla_{\alpha} \mathcal{L}_{val}\left(w^{\prime}, \alpha\right) - \xi \frac{\nabla_{\alpha} \mathcal{L}_{tr}\left(w_{val}^{+}, \alpha \right)-\nabla_{\alpha} \mathcal{L}_{tr}\left(w_{val}^{-}, \alpha \right)}{2 \epsilon^{val}} \right)  \right.\\ \nonumber
&\left. +\lambda \left( \nabla_{\alpha} \mathcal{L}_{tr}\left(w^{\prime}, \alpha\right) - \xi \frac{\nabla_{\alpha} \mathcal{L}_{tr}\left(w_{tr}^{+}, \alpha\right)-\nabla_{\alpha} \mathcal{L}_{tr}\left(w_{tr}^{-}, \alpha \right)}{2 \epsilon^{tr}} \right) \right]
\end{aligned}    
\end{equation}
where $w^{\prime}=w-\xi \nabla_{w} \mathcal{L}_{\mathrm{tr}}(w, \alpha)$,  $w_{\mathrm{val}}^{ \pm}=w \pm \epsilon^{val} \nabla_{w^{\prime}} \mathcal{L}_{v a l}\left(w^{\prime}, \alpha\right)$, $w_{tr}^{ \pm}=w \pm \epsilon^{tr} \nabla_{w^{\prime}} \mathcal{L}_{tr}\left(w^{\prime}, \alpha\right)$.  $\epsilon^{tr}$ and $\epsilon^{val}$ are two scalars.
More details on the derivation of the \mldas\ second-order method are placed into the Appendix.

\begin{figure*}[ht!]
    \centering
    \includegraphics[width=0.9\textwidth]{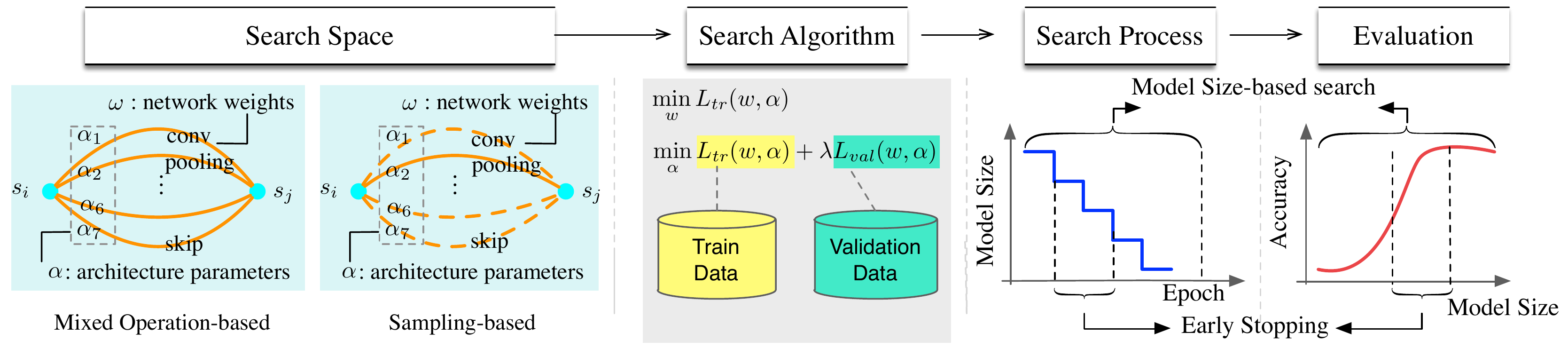}
    \caption{The Overview of the \mldas\ Framework.}
    \label{fig:framework}
\end{figure*}
Another benefit of mixed-level optimization is that it can embed more information. 
In fact, when updating $\alpha$, the training loss can also effectively judge how well the neural network structure performs. Thus, it is better to fully exploit the information embedded in both the training and validation loss when updating $\alpha$.

Next, we will analyze the benefits of mixed-level reformulation and conclude that the \mldas-1st method is a better choice in solving the NAS problem.

\subsection{Comparison between MiLeNAS and DARTS}
\label{sec:comparison-with-bilevel}
\paragraph{\mldas \text{-1st} v.s. DARTS-2nd} As we discussed,  mixed-level optimization avoids the overfitting issue and fully exploits training and validation data. Although DARTS-2nd also incorporates the training data, compared to \mldas \text{-1st}, it has gradient deviation and searches inefficiently due to gradient approximation. 
To be more specific, when optimizing $\alpha$ in bilevel optimization (Equation \ref{eq:bi_prob}), DARTS-2nd \cite{liu2018darts} approximates $w$ with one step update: $ \nabla_{\alpha} \mathcal{L}_{v a l}\left(w^{*}(\alpha), \alpha\right) \approx \nabla_{\alpha} \mathcal{L}_{v a l}\left(w-\xi \nabla_{w} \mathcal{L}_{t r a i n}(w, \alpha), \alpha\right)$, and then applies the chain rule to yield:

\begin{equation}
\small
    \begin{aligned}
    \nabla_{\alpha} \mmL_{\mathrm{val}}\left(w^{*}(\alpha), \alpha\right) &\approx \underbrace{\nabla_{\alpha} \mathcal{L}_{v a l}\left(w^{\prime}, \alpha\right)}_{g_1}- \\
&\underbrace{\xi \nabla_{\alpha, w}^{2} \mathcal{L}_{t r a i n}(w, \alpha) \nabla_{w^{\prime}} \mathcal{L}_{v a l}\left(w^{\prime}, \alpha\right)}_{g_2},  
    \end{aligned}
\label{eq:chain-rule}
\end{equation}
where $w^{\prime}=w-\xi \nabla_{w} \mathcal{L}_{t r a i n}(w, \alpha)$ denotes the weights for a one-step forward model.
To avoid an expensive matrix-vector product in its second term $g_2$, DARTS-2nd uses the finite difference approximation to reduce its complexity:
\begin{equation}
\small
    \begin{aligned}
    \nabla_{\alpha} \mmL_{\mathrm{val}}&\left(w^{*}(\alpha), \alpha\right) \approx \nabla_{\alpha} \mathcal{L}_{v a l}\left(w^{\prime}, \alpha\right) - \\
&\xi \frac{\nabla_{\alpha} \mathcal{L}_{t r a i n}\left(w_{val}^{+}, \alpha\right)-\nabla_{\alpha} \mathcal{L}_{t r a i n}\left(w_{val}^{-}, \alpha\right)}{2 \epsilon^{val}} 
    \end{aligned}
\label{eq:val}
\end{equation}
where $w_{val}^{ \pm}=w \pm \epsilon^{val} \nabla_{w^{\prime}} \mathcal{L}_{v a l}\left(w^{\prime}, \alpha\right)$.

This brings up two problems: 1. We can clearly see from Equation \ref{eq:val} that the second-order approximation has a superposition effect: the second-order approximation $\alpha$ is built upon the one-step approximation of $w$. This superposition effect causes gradient error, leading to deviation from the true gradient. Consequently, this gradient error may lead to an unreliable search and sub-optimal architectures; 2. Equation \ref{eq:val} requires two forward passes for the weights $w$ and two backward passes for $\alpha$, which is inefficient.

In contrast, our \mldas \text{-1st} method only uses the first-order information (shown in Equation \ref{eq:mixed_level}), which does not involve gradient errors caused by superposition approximation. 
Furthermore, comparing Equation \ref{eq:val} with our update of $\alpha$ in Equation \ref{eq:mixed_level}, we can see that our \mldas \text{-1st} requires far fewer operations, resulting in a faster convergence speed. Our experiments concur this analysis.

\paragraph{\mldas \text{-1st} v.s. DARTS-1st}
DARTS also proposes to use the first-order algorithm to solve the bilevel optimization, which can be summarized as
\begin{align*}
    &w = w - \eta_w \nabla_w \mmL_{\mathrm{tr}}(w, \alpha), \\
    &\alpha = \alpha - \eta_\alpha \nabla_\alpha \mmL_{\mathrm{val}}(w,\alpha).
\end{align*}

Although both \mldas \text{-1st} and DARTS-1st share a simple form, they have fundamental differences. When updating $\alpha$, \mldas-1st (\eqref{eq:mixed_level}) takes advantage of both the training and validation losses and obtains a balance between them by setting parameter $\lambda$ properly, while DARTS-1st only exploits information in validation loss $\mmL_{\mathrm{val}}(w,\alpha)$. Therefore, \mldas-1st achieves better performance than DARTS-1st. 
Moreover, the experiments in the original DARTS paper show that DARTS-2nd outperforms DARTS-1st since DARTS-2nd also utilizes the training loss when updating $\alpha$ (refer to \eqref{eq:val}).
Thus, this also provides evidence that exploiting more information (from the training dataset) can help \mldas\ to obtain a better performance than DARTS-1st.

\paragraph{\mldas \text{-1st} v.s. \mldas \text{-2nd}} To fully understand \mldas, we further investigate the effectiveness of \mldas-2nd. Our experiments show that \mldas \text{-2nd} is not as good as \mldas-1st. Compared to \mldas \text{-1st}, its searched architecture has a lower accuracy, and its search speed is slow. This conclusion supports our expectations because \mldas-2nd and DARTS-2nd both have the same gradient error issue in which the second-order approximation of the true gradient causes a large deviation in the gradient descent process. This approximation only brings negative effects since MiLeNAS-1st already fully exploits the information embedded in the training and validation losses. 
Thus, in practice, we conclude that among these methods, \mldas-1st could be the first choice in solving the NAS problem. 
More experimental details are covered in the appendix.

In summary, our method not only is simple and efficient, but also avoids the gradient error caused by the approximation in the bilevel second-order method. Thus, it can search with more stability and a faster speed and find a better architecture with higher accuracy.

\subsection{Beyond the DARTS Framework}
\label{sec:beyond}
Motivated by the above analysis and experimental results, \mldas\ further upgrades the general DARTS framework. As shown in Figure \ref{fig:framework}, there are three key differences.

\paragraph{MiLeNAS on Gradient-based Search Spaces}
First, since our proposed \mldas\ is a generic framework for gradient-based NAS, we evaluate our method in two search space settings. The first is the mixed-operation search space defined in DARTS, where architecture search only performs on convolutional cells to find candidate operations (e.g., convolution, max pooling, skip connection, and zero) between nodes inside a cell. To make the search space continuous, we relax the categorical choice of a connection to a softmax over all possible operations:
\begin{equation}
\bar{o}^{(i,j)} (x) = \sum_{k=1}^{d} \underbrace{\frac{\exp(\alpha_k^{(i,j)})}{\sum_{k'=1}^{d} \exp(\alpha_{k'}^{(i,j)})}}_{p_k} o_k(x).
\label{equ: mixed operations}
\end{equation}
The weight $p_k$ of the mixed operation $\bar{o}^{(i,j)} (x)$ for a pair of nodes $(i,j)$ is parameterized by a vector $\alpha^{i,j}$.
Thus, all architecture operation options inside a network (model) can be parameterized as $\alpha$. By this definition, \mldas\ aims at simultaneously optimizing architecture parameters $\alpha$ and model weights $w$. 

Another is the sampling search space: instead of the mixed operation as Equation \ref{equ: mixed operations}, GDAS \cite{dong2019searching} uses a differentiable sampler (Gumbel-Softmax) to choose an operation between two nodes in a cell:
\begin{equation}
    \Tilde{p}_k^{(i,j)} (x) =  \frac{\exp((\alpha_k^{(i,j)} + u_k)/\tau) }{\sum_{k'=1}^{d} \exp((\alpha_{k'}^{(i,j)}+ u_k)/\tau)},
\end{equation}
where $u_k$ are  i.i.d samples drawn from the Gumbel$(0,1)$ distribution and $\tau$ is the softmax temperature. We substitute bilevel optimization in GDAS with mixed-level optimization to verify the versatility of \mldas.

In fact, we can design any search space using the \mldas\ framework. In this paper, we demonstrate mixed-level optimization using DARTS and GDAS.

\paragraph{Model Size-based Searching} We propose the model size-based searching, which is defined as searching optimal architectures in different model sizes in a single run. To be more specific, during the search, we track the model size and its best validation accuracy after every epoch, then evaluate the performance of the optimal architecture in each model size. The advantage is that we can get multiple architectures with different parameter sizes with only a single run. Our motivations are as follows: 1) to fully understand the search process with different optimization methods, we use model size-based search and find that \mldas\ is more reliable in the search process: it stably acts in a regular model size evolution pattern (will be introduced in Section \ref{sec:strategy}); 2) we hypothesize that a good NAS search method can fully exploit the accuracy in different model sizes, meaning that in the search process, the architecture with the highest validation accuracy in each model size is expected to perform excellently after architecture evaluation. This is largely ignored by previous NAS methods. In Section \ref{sec:strategy}, we present experimental results of this search strategy and provide some insights for neural architecture design.

\paragraph{Early Stopping Strategy} Early stopping strategy is motivated by the observation of the search process when using model size-based search. We find that after a certain number of epochs (around 25 epochs in DARTS and \mldas), the model size will decrease. Since we know that larger model sizes may lead to better performance, we stop searching if the model size is less than the expected size. Through our experimental analysis, by drawing the relationship between the model size and the model performance (accuracy), we can determine the best stopping timing during the search process (will be introduced in Section \ref{sec:strategy}).

With the improvements discussed above, we summarize the \mldas\ framework as Algorithm \ref{alg:mixed-level}.
\begin{algorithm}[htb]
    \caption{\mldas\ Algorithm}
    \begin{small}
        \begin{algorithmic}[1]
            \STATE Define the search space;
            \WHILE{not converge} 
            \FOR{$e$ in epoch}
            \FOR{minibatch in training and validation data}
            \STATE $w = w - \eta_w \nabla_w \mmL_{\mathrm{tr}}(w, \alpha);$
            \STATE $\alpha = \alpha - \eta_\alpha \left(\nabla_\alpha  \mmL_{\mathrm{tr}}(w, \alpha) +  \lambda \nabla_\alpha \mmL_{\mathrm{val}} (w, \alpha)\right);$
            \ENDFOR
            \STATE Save the optimal structures under different model sizes;
            \IF{current model size is less than the expected size}
            \STATE break;
            \ENDIF
            \ENDFOR
            \ENDWHILE
            \STATE Evaluate on the searched neural network architecture.
        \end{algorithmic}
    \end{small}
\label{alg:mixed-level}
\end{algorithm}

\section{Experiments and Results}
\begin{figure*}[htb]
\centering
    \begin{subfigure}{0.32\linewidth}
        \centering
        \includegraphics[width=\linewidth]{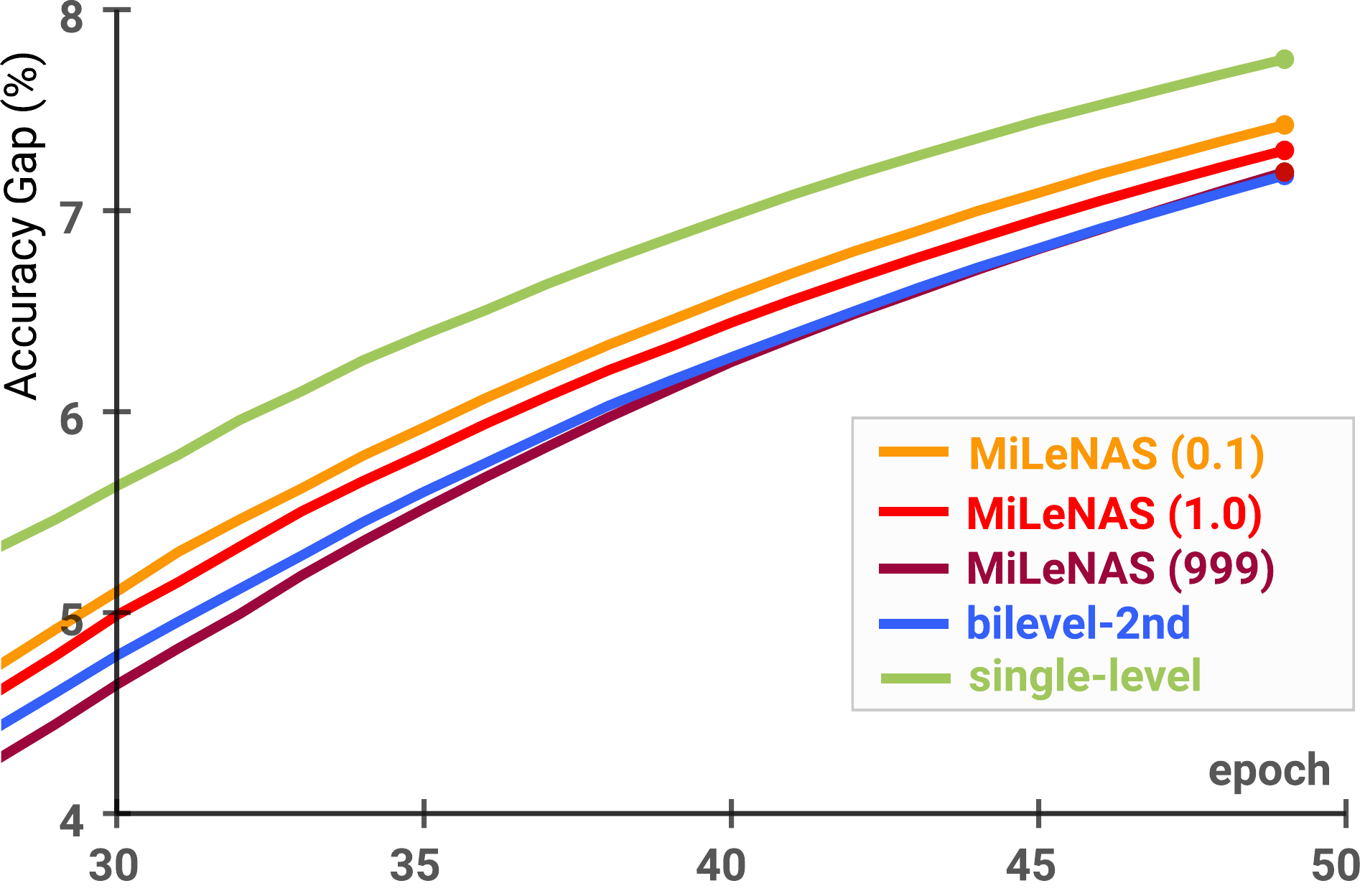}
        \caption{Searching Training and Validation Gap}
        \label{fig:correlation}
    \end{subfigure}
    \begin{subfigure}{0.32\linewidth}
        \centering
        \includegraphics[width=\linewidth]{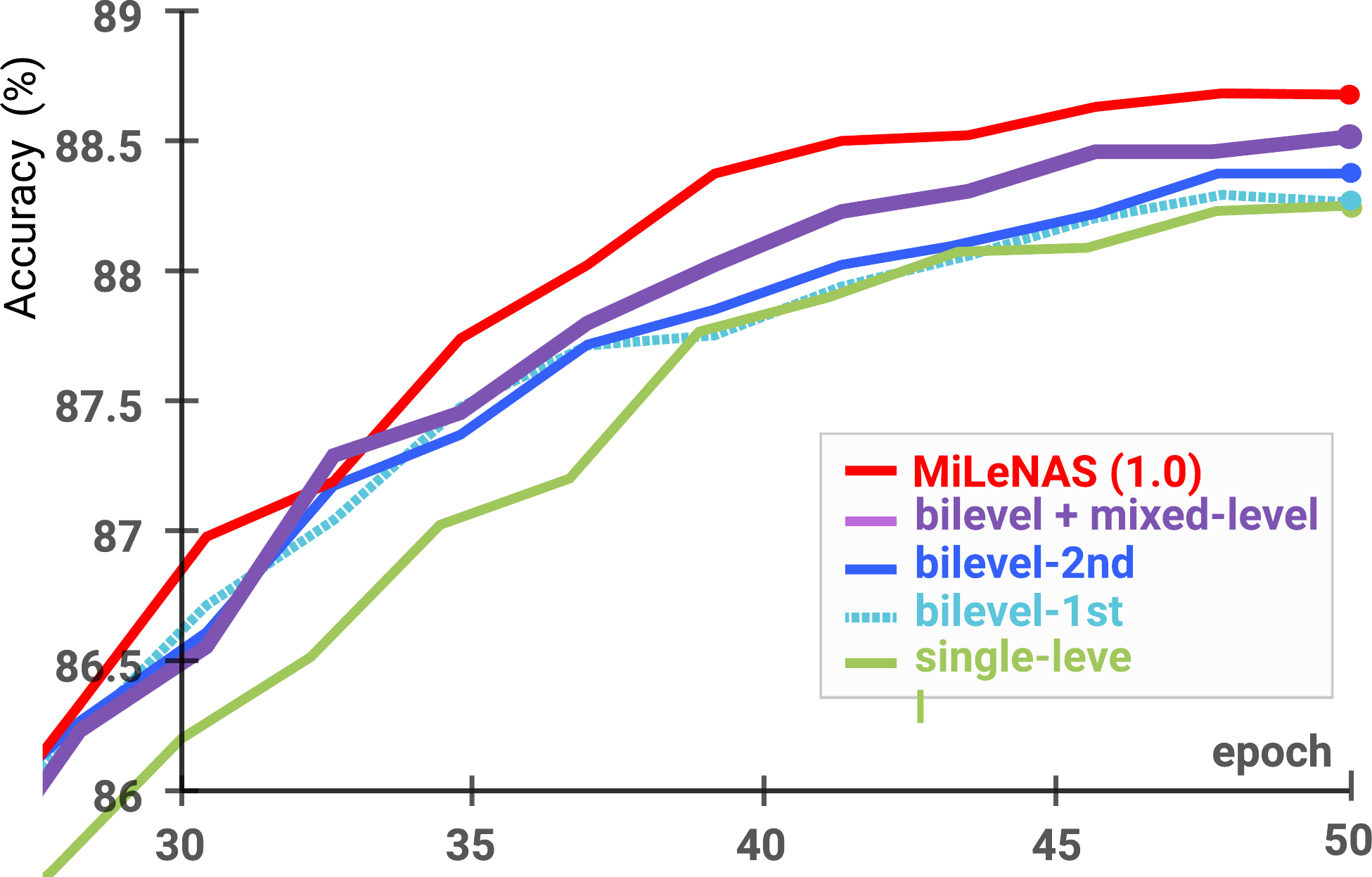}
        \caption{Searching Validation Accuracy}
        \label{fig:validation_acc}
    \end{subfigure}
    \begin{subfigure}{0.32\linewidth}
        \centering
        \includegraphics[width=\linewidth]{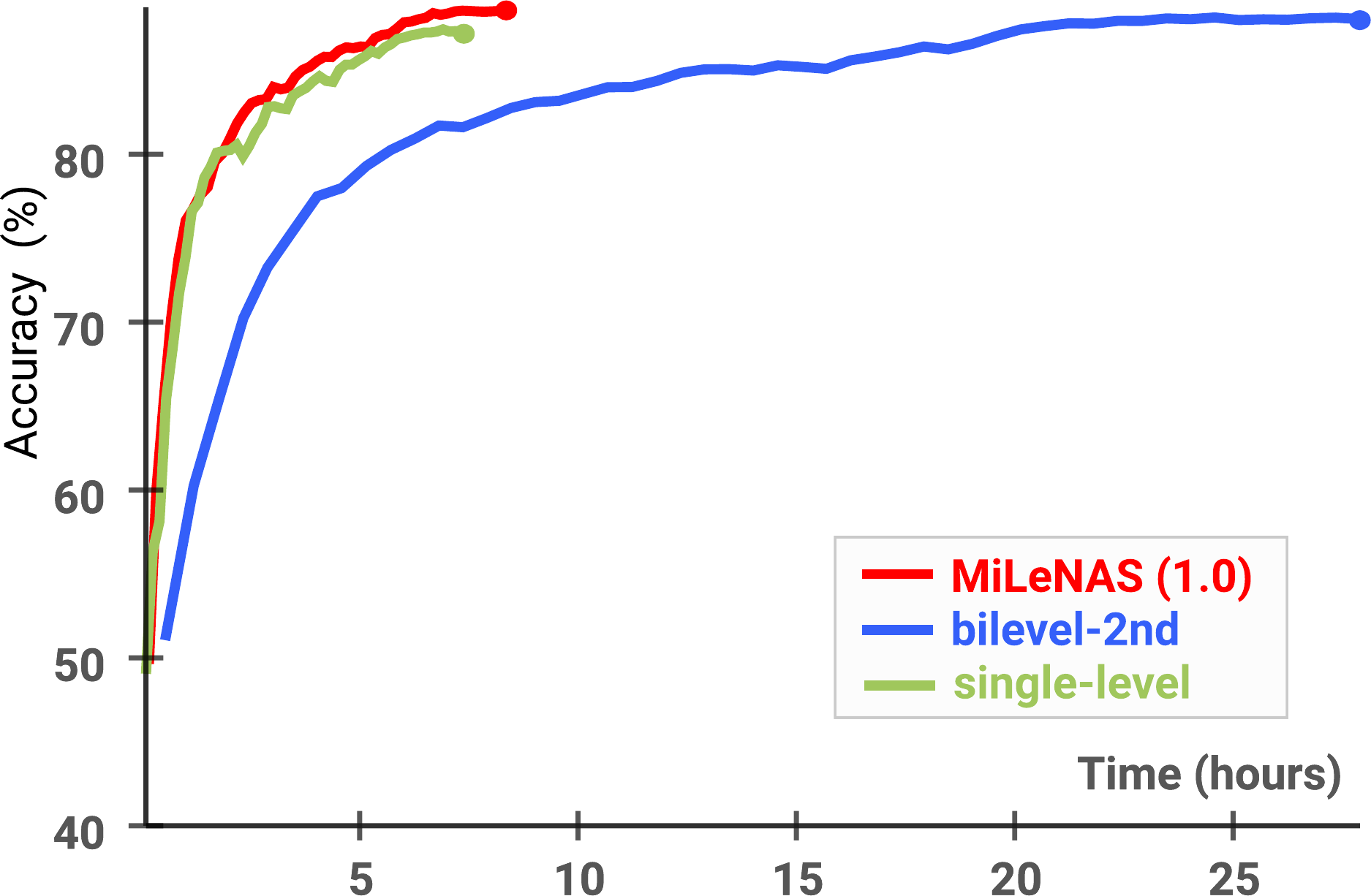}
        \caption{Searching Time}
        \label{fig:training_time}
    \end{subfigure}
    \caption{Comparing \mldas-1st with single-level and bilevel methods. }
    \label{fig:Searching}
\end{figure*}



\subsection{Settings}
\mldas\ contains two stages: architecture search and architecture evaluation. The image classification dataset CIFAR-10 \cite{krizhevsky2009learning} is used for the search and evaluation, while the ImageNet dataset is used for the transferability verification. To maintain a fair comparison, a search space definition similar to that of DARTS was chosen. In the search stage, the validation dataset is separated from the training dataset, and each method is run four times.
In the evaluation stage, the architecture with the highest validation accuracy in the search stage is chosen. Our code implementation is based on PyTorch 1.2.0 and Python 3.7.4. All experiments were run on NVIDIA Tesla V100 16GB.
Hyperparameter settings are kept the same as DARTS. More Details regarding the experimental settings are presented in Appendix. 

\subsection{Comparison with Single-level and Bilevel methods}
Our intensive experimental evidence demonstrates that the  architecture which has the highest validation accuracy during searching also has a larger probability to obtain the highest accuracy in the evaluation stage. Thus, to demonstrate the advantage of MiLeNAS, we first compare the validation accuracy of different methods in the search stage.

We correlate our \mldas-1st method with single-level and bilevel methods by verifying the gap between training accuracy and validation accuracy to measure the overfitting with respect to the structure parameter $\alpha$. 
For \mldas, we choose three $\lambda$ settings ($\lambda = 0.1, \lambda = 1, \lambda = 999$) to assign different proportions between the training loss and validation loss. For the single-level method, we update both $\alpha$ and $w$ on the training dataset, while for the bilevel method, we use DARTS-2nd \cite{liu2018darts}. 
The epoch number is set to 50 (as set in second-order DARTS). 
In total, the five settings are run four times each, and the final results are based on the averages. 
From the results shown in Figure \ref{fig:correlation}, we see that the single-level method has the largest gap, while the bilevel method has the smallest gap. Our mixed-level method lies between them: when $\lambda$ is small (0.1), the gap is closer to that of the single-level method, while when $\lambda$ is large (999), the gap is closer to that of the bilevel method. 
Thus, this result confirms our assertion that
single-level and bilevel optimizations are special cases of mixed-level optimization with $\lambda = 0$ and $\lambda \to\infty$, respectively.

As demonstrated in Figure \ref{fig:validation_acc}, \mldas\ with  $\lambda=1$ achieves the highest validation accuracy. The validation accuracy of DARTS-2nd is larger than DARTS-1st (labeled as \textit{bilevel-1st} in Figure \ref{fig:validation_acc}), which is the same result as in the original DARTS paper. The single-level method gains the lowest validation accuracy. This comparison of the validation accuracy is aligned with our theoretical analysis in Section \ref{sec:comparison-with-bilevel}: \mldas\ is not only simple and efficient but also avoids the gradient error caused by the approximation in the bilevel second-order method. 

To further confirm the effectiveness of \mldas, we perform another experiment running bilevel optimization for the first 35 epochs, then switching to our \mldas\ method. When comparing its result (the bold purple curve in Figure \ref{fig:validation_acc}) to that of the pure bilevel optimization (the blue curve in Figure \ref{fig:validation_acc}), we see that \mldas\ continues to improve the validation accuracy in the late phase of the search process. This observation confirms that our mixed-level algorithm can mitigate the gradient approximation issue, outperforming bilevel optimization.

Furthermore, as shown in Figure \ref{fig:training_time}, \mldas\ is over three times faster than DARTS-2nd. \mldas\ performs a faster search due to its simple first-order algorithm, while the second-order approximation in DARTS requires more gradient computation (discussed in section \ref{sec:comparison-with-bilevel}). 

\subsection{Evaluation Results on CIFAR-10}
\label{sec:results}

\begin{table*}[hbt]
\footnotesize
\centering
\caption{Comparison with state-of-the-art image classifiers on CIFAR-10.}
\label{table:comparison-CIFAR10}
\begin{threeparttable}
\centering

\begin{tabular}{lc c c c c}
    \toprule

\textbf{Architecture} & \textbf{Test Error (\%)} & \textbf{Params (M)} & \textbf{Search Cost (GPU days)} & \textbf{Search Method} \\
\midrule
DenseNet-BC \cite{huang2017densely} & 3.46 & 25.6 & - & manual\\ 

NASNet-A + cutout \cite{zoph2018learning} & 2.65 & 3.3 & 2000 & RL\\ 


BlockQNN \cite{zhong2018blockqnn} & 3.54 & 39.8 & 96 & RL\\ 


AmoebaNet-B + cutout \cite{real2019regularized} & $2.55\pm0.05$ & 2.8 & 3150 & evolution\\ 
Hierarchical evolution \cite{liu2017hierarchical} & $3.75\pm0.12$ & 15.7 & 300 & evolution\\ 
PNAS \cite{liu2018progressive} & $3.41\pm0.09$ & 3.2 & 225 & SMBO\\ 

ENAS + cutout \cite{pham2018efficient}\tnote{$\dagger$} & 2.89 & 4.6 & 0.5 & RL\\

\midrule


DARTS (second order) \cite{liu2018darts} & $2.76\pm0.09$ & 3.3 & 1 & gradient-based\\
SNAS (moderate) \cite{xie2018snas} & $2.85\pm 0.02$ & 2.8 & 1.5 & gradient-based\\
SNAS (aggressive) \cite{xie2018snas} & $3.10\pm0.04$ & 2.3 & 1.5 & gradient-based\\
GDAS \cite{dong2019searching} & 2.82 & 2.5 & 0.17 & gradient-based\\
\midrule
\textbf{\mldas}\tnote{*} & \textbf{2.51$\pm$ 0.11 (best: 2.34)} & \textbf{3.87} & 0.3 & gradient-based\\
 
\textbf{\mldas}\tnote{*} & \textbf{2.80$\pm$ 0.04 (best: 2.72)} & \textbf{2.87} & 0.3 & gradient-based\\

\mldas\tnote{*} & \textbf{2.50} & \textbf{2.86} & 0.3 & gradient-based\\

\mldas\tnote{*} & \textbf{2.76} & \textbf{2.09} & 0.3 & gradient-based\\
\bottomrule 
\end{tabular}
  \begin{tablenotes}
      \small
      \item[*] We get multiple results by using model size-based searching (introduced in Section \ref{sec:strategy}); the search time is calculated without the early stopping strategy (around 8 hours). If the early stopping strategy is used, the search cost can further be reduced to around 5 hours.
    \end{tablenotes}
    \end{threeparttable}
\end{table*}
In the evaluation stage, 20 searched cells are stacked to form a larger network, which is subsequently trained from scratch for 600 epochs with a batch size of 96 and a learning rate set to 0.025. For fair comparison, every architecture shares the same hyperparameters as the DARTS bilevel method.
The CIFAR-10 evaluation results are shown in Table \ref{table:comparison-CIFAR10} (all architectures are searched using $\lambda=1$). The test error of our method is on par with the state-of-the-art RL-based and evolution-based NAS while using three orders of magnitude fewer computation resources. 
Furthermore, our method outperforms ENAS, DARTS-2nd, SNAS, and GDAS with both a lower error rate and fewer parameters. 
We also demonstrate that our algorithm can search architectures with fewer parameters while maintaining high accuracy.

\begin{table*}[htb!]
\footnotesize
\caption{Comparison with state-of-the-art image classifiers on ImageNet.}
\label{table:comparison-ImageNet}
\centering
\begin{threeparttable}
\centering
  
\begin{tabular}{lc c c c c c p{30pt}}
\toprule
\centering
\multirow{2}*{\textbf{Architecture}} & \multicolumn{2}{c}{\textbf{Test Error (\%)}} & \multirow{2}*{\textbf{Params (M)}}& \multirow{2}*{\textbf{$+\times$ (M)}}& \multirow{2}*{\textbf{Search Cost (GPU days)}} & \multirow{2}*{\textbf{Search Method}} \\
\cmidrule{2-3}
 & top-1 & top-5 &  &  &  &  \\
\midrule
Inception-v1 \cite{szegedy2015going} & 30.2 & 10.1 & 6.6 & 1448 & - & manual\\ 

MobileNet \cite{howard2017mobilenets} & 29.4 & 10.5 & 4.2 & 569 & - & manual\\ 

ShuffleNet \cite{zhang2018shufflenet} & 26.3 & - & $\sim5$ & 524 & - & manual\\ 

NASNet-A \cite{zoph2018learning} & 26.0 & 8.4 & 5.3 & 564 & 2000 & RL\\ 



 AmoebaNet-A \cite{real2019regularized} & 25.5 & 8.0 & 5.1 & 555 & 3150 & evolution\\ 


AmoebaNet-C \cite{real2019regularized} & 24.3 & 7.6 & 6.4 & 570 & 3150 & evolution\\ 

PNAS \cite{liu2018progressive} & 25.8 & 8.1 & 5.1 & 588 & $\sim225$ & SMBO\\ 
\midrule
DARTS \cite{liu2018darts} & 26.7 & 8.7 & 4.7 & 574 & 1 & gradient-based\\ 

SNAS \cite{xie2018snas} & 27.3 & 9.2 & 4.2 & 522 & 1.5 & gradient-based\\ 
GDAS \cite{dong2019searching} & 27.5 & 9.1 & 4.4 & 497 & 0.17 & gradient-based\\ 
GDAS \cite{dong2019searching} & 26.0 & 8.5 & 5.3 & 581 & 0.21 & gradient-based\\ 
\midrule
\mldas\tnote{*} & \textbf{25.4} & \textbf{7.9} & \textbf{4.9} & \textbf{570} & \textbf{0.3} & gradient-based\\ 
\mldas\tnote{*} & \textbf{24.7} & \textbf{7.6} & \textbf{5.3} & \textbf{584} & \textbf{0.3} & gradient-based\\
\bottomrule
  \end{tabular}
  \begin{tablenotes}
      \small
      \item[*] We gain multiple architectures by using model size-based searching (introduced in Section \ref{sec:strategy}), and then do transfer learning on ImageNet.
    \end{tablenotes}
    \end{threeparttable}
\end{table*}

\subsection{Transferability on ImageNet}

Transferability is a crucial criterion used to evaluate the potential of the learned cells \cite{zoph2018learning}. To show if the cells learned through our method on CIFAR-10 can be generalized to larger datasets, we use the same cells as in CIFAR-10 for the classification task on ImageNet. 
Table \ref{table:comparison-ImageNet} presents the results of the evaluation on ImageNet and shows that the cells found by our method on CIFAR-10 can be successfully transferred to ImageNet.
Our method can find smaller cell architectures that achieve a relatively better performance at speeds three times faster than the bi-level method (DARTS-2nd). Hyperparameter Settings are presented in Appendix.
\section{Beyond the DARTS Framework}
In this section, we demonstrate the effectiveness of \mldas\ in other NAS frameworks, and then propose two strategies: model sized-based searching and early stopping.

\begin{figure*}[hbt!]
    \centering
    ~ 
      \begin{subfigure}[b]{0.3\linewidth}
        \includegraphics[width=\linewidth]{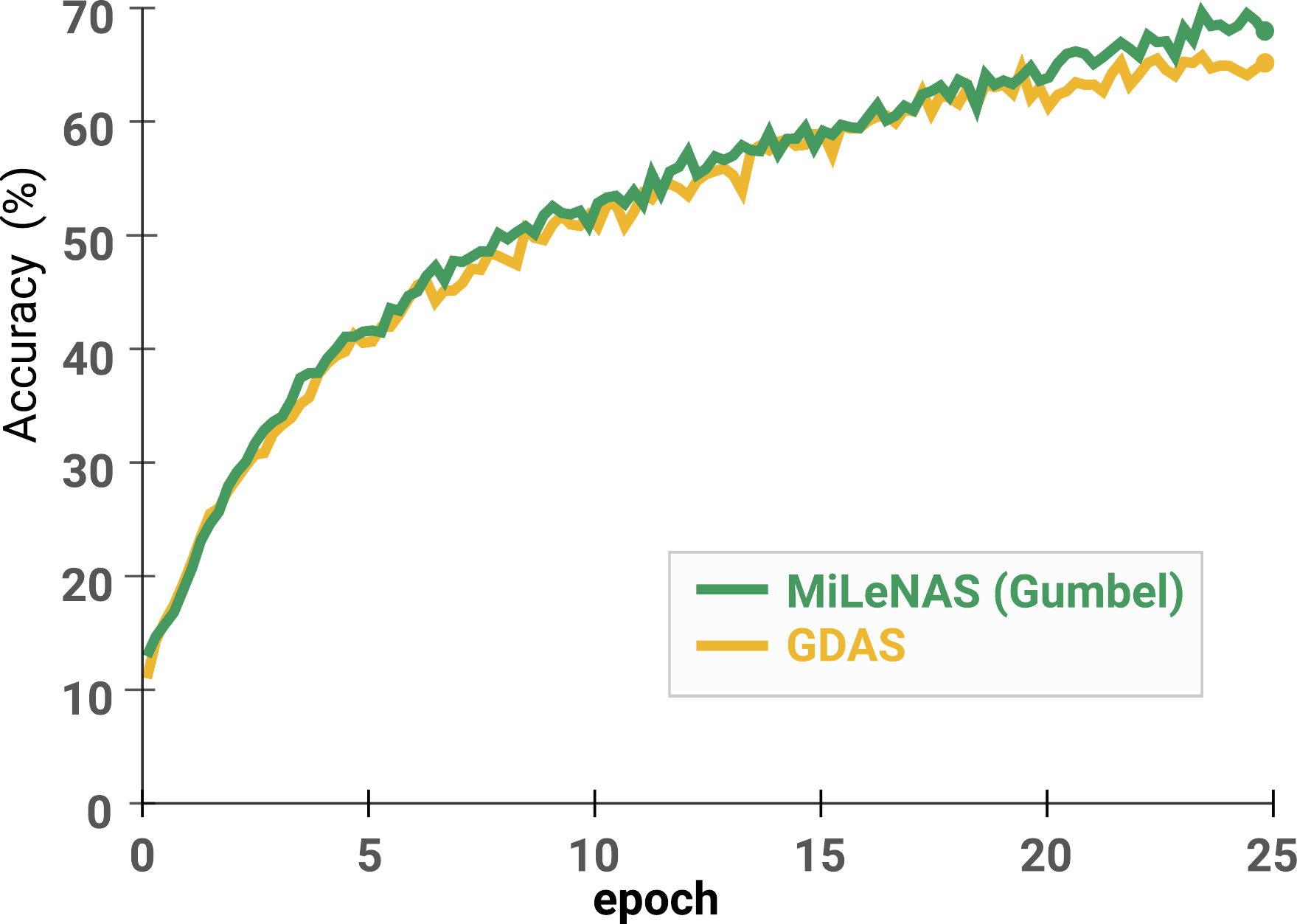}
        \caption{Applying MiLeNAS on GDAS}
        \label{fig:GDAS}
      \end{subfigure}
    \begin{subfigure}[b]{0.32\linewidth}
        \includegraphics[width=\linewidth]{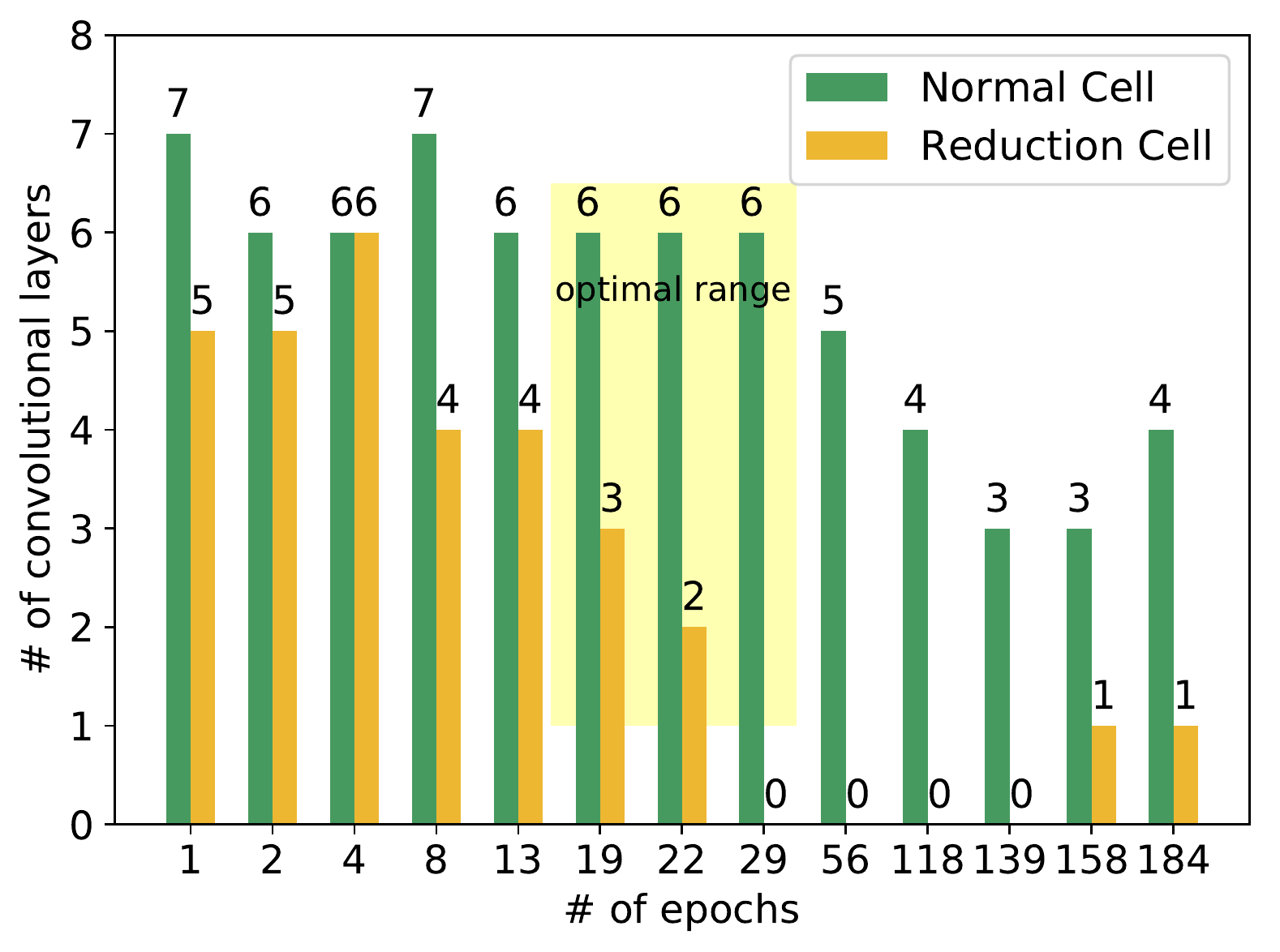}
        \caption{Model Size-based Searching Process}
        \label{fig:params-valid-1}
    \end{subfigure}
    \begin{subfigure}[b]{0.32\linewidth}
        \includegraphics[width=\linewidth]{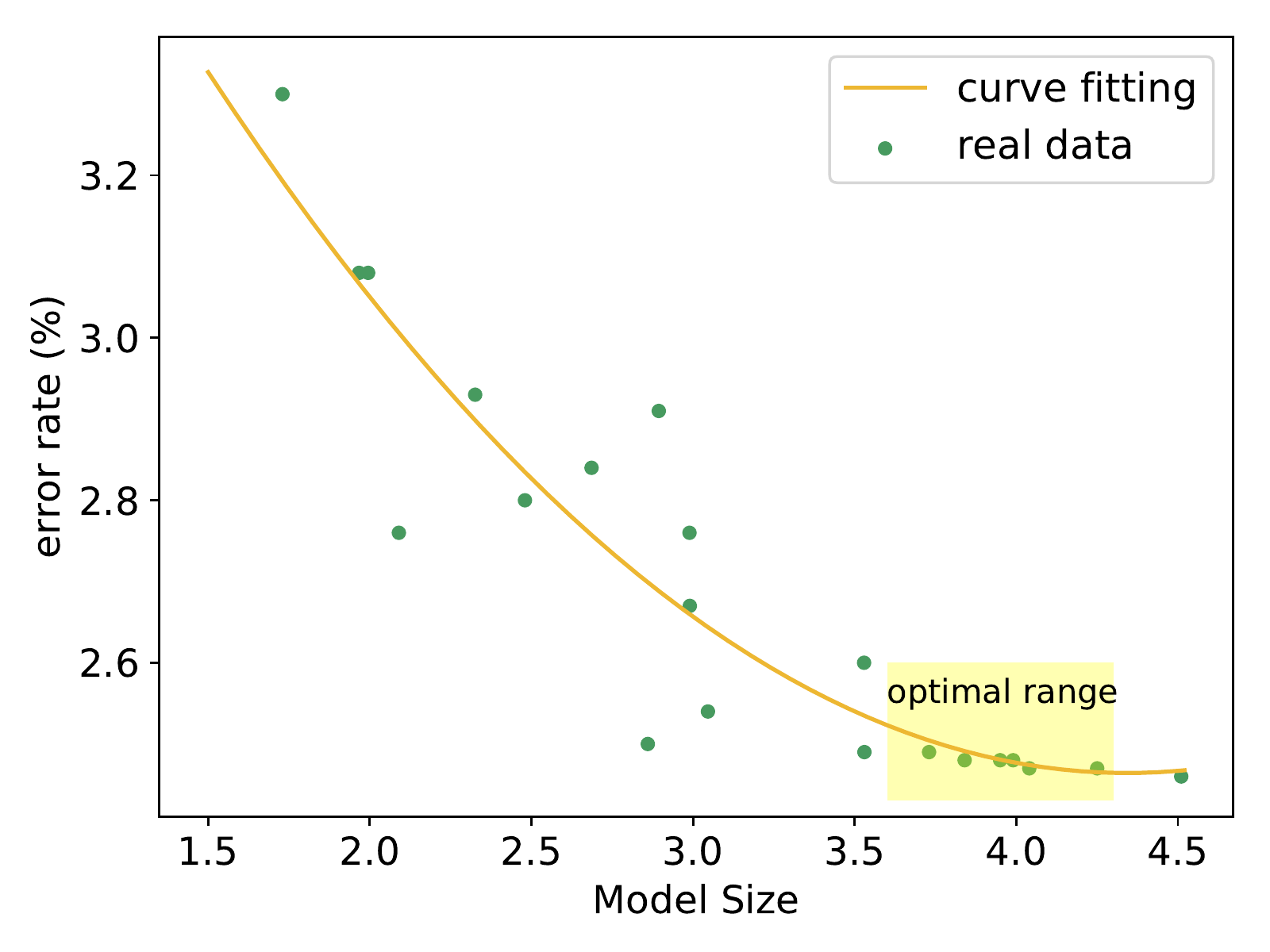}
        \caption{Error Rate over Varying Model Size}
        \label{fig:params-valid-2}
    \end{subfigure}
    \caption{Evaluation on the Searched Architecture with Different Model Sizes}
\label{fig:params-valid}
\end{figure*}

\mldas\ is universal and can be used as a substitute for the bilevel optimization in other NAS methods to improve their search performances. We perform verification experiments on the Gumbel-Softmax sampling method GDAS \cite{dong2019searching}. We reproduce GDAS\footnote{As of the publication of this paper, GDAS still has not published the source code.}  and substitute its bilevel optimization with \mldas, denoted as \mldas \space(Gumbel). As shown in Figure \ref{fig:GDAS}, \mldas \space(Gumbel) can achieve a better validation accuracy (GDAS: 65.79\%; \mldas \space(Gumbel): \textit{69.56\%}), leading to better architectures with lower error rates (GDAS: 2.82\%; \mldas \space(Gumbel): \textit{2.57\%}).

\label{sec:strategy}
\subsection{Model Size-based Searching}

\textbf{Model Size Tracking}. To understand the model size evolution during the searching process, for the architecture searched in each epoch, we track the best validation accuracy for different model sizes, which is calculated by counting the number of convolution operations in the searched cell. We track the model size in this way because different discrete operation choices (determined by $\alpha$) in a cell determine the model size (e.g., the model size of an architecture with more convection operations is larger than an architecture with more skip-connection operations).

\textbf{Observations}. By tracking the model size, we find that \mldas\ has an obvious phase-by-phase optimization characteristic during the search process. As illustrated in Figure \ref{fig:params-valid-1}, each phase optimizes the network architecture under a certain range of the model size (counted by the convolutional layer number) and then reduces the model size before entering another optimization phase. We evaluate the optimal architecture in each phase and find the relationship between the model size and the model performance (accuracy). From Figure \ref{fig:params-valid-2}, we learn that when the model size increases,  the model performance also increases.  However, this growth reaches its limit between the optimal range (between 3.5M and 4.5M). Subsequently, the model performance is unable to improve even with an increase in the parameter number. 

Our experiment on DARTS does not consistently show the same model size decreasing characteristics as in \mldas. In other words, to summarize the model size and accuracy relation, we must run much more search rounds in DARTS since it does not have a stable pattern. We argue that this regular pattern seen in \mldas\ is attributed to our mixed-level optimization since it does not suffer from gradient error by using second-order approximation. 

\textbf{Insight}. The above observation during the search process drives our search strategy design. We define the model size-based searching as completing the search process in a large number of epochs with model size tracking, and then evaluating the network architecture accuracy under different model sizes. This provides three potential benefits for neural architecture design: 1) for a specific learning task, the most economical neural network architecture must be examined, and the redundant parameter quantity cannot bring additional benefits; 2) this method has the potential to become an alternative method for model compression since it can find multiple optimal architectures under different computational complexities; 3) most importantly, we may figure out a regular pattern between the parameter number and the architecture accuracy, allowing us to refine a strategy further to expedite the searching process. In our case, we have found that the early stopping strategy can remarkably accelerate the search speed.

\subsection{Early Stopping Strategy} 
The timing of stopping the search is inspired by the fact that the optimal range of parameter numbers in Figure \ref{fig:params-valid-2} (highlighted by yellow square) is found in the early phase of the searching process in Figure \ref{fig:params-valid-1} (highlighted by yellow square). For example, we stop searching when the parameter number reaches $6$ convolution operations in the normal cell and $0$ convolution operation in the reduction cell. 
When utilizing this stopping strategy, searching for the optimal architecture on CIFAR-10 with \mldas\ only costs around 5 hours.



\section{Conclusion}
We proposed \mldas, a novel perspective to the NAS problem, and reformulated it as mixed-level optimization instead of bilevel optimization. \mldas\ can alleviate gradient error caused by approximation in bilevel optimization and benefits from the first-order efficiency seen in single-level methods. Thus, \mldas\ can search for better architectures with a faster convergence rate. The extensive experiments on image classification have demonstrated that \mldas\ can gain a lower validation error at a search time three times shorter than 2nd-order bilevel optimization. \mldas\ is a generic method. Its applicability experiments verify that it can be used in the sampling-based method to search for better architectures. 
Model size-based search and early stopping strategies further speed up the searching process and additionally provide several insights into neural architecture design as well.

\small{
\bibliography{milenas_review}
\bibliographystyle{milenas_review}
}

\clearpage





\appendix
\section{Experiment Details}
\subsection{Search Space Definition}
\label{sec:search-space}

We adopt the following 8 operations in our CIFAR-10 experiments: 3 $\times$ 3 and 5 $\times$ 5 separable convolutions, 3 $\times$ 3 and $5 \times 5$ dilated separable convolutions, 3 $\times$ 3 max pooling, 3 $\times$ 3 average pooling, identity, and zero.

The network is formed by stacking convolutional cells multiple times. Cell $k$ takes the outputs of cell $k-2$ and cell $k-1$ as its input. Each cell contains seven nodes: two input nodes, one output node, and the other four intermediate nodes inside the cell. The input of the first intermediate node is set equal to two input nodes, and the other intermediate nodes take all previous intermediate nodes' output as input. The output node concatenates all intermediate nodes' outputs in depth-wise. There are two types of cells: the normal cell and the reduction cell. The reduction cell is designed to reduce the spatial resolution of feature maps, locating at the 1/3 and 2/3 of the total depth of the network. Architecture parameters determine the discrete operation value between two nodes. All normal cells and all reduction cells share the same architecture parameters $\boldsymbol{\alpha}_{\text {n}}$ and $\boldsymbol{\alpha}_{\text {r}}$, respectively. By this definition, our method alternatively optimizes architecture parameters ($\boldsymbol{\alpha}_{\text {n}}$, $\boldsymbol{\alpha}_{\text {r}}$) and model weight parameters $\boldsymbol{w}$. 

\subsection{Transferability on ImageNet}
\label{apd:imagenet}
The model is restricted to be less than 600M. A network of 14 cells is trained for 250 epochs with a batch size of 128, weight decay $3 \times 10^{-5}$, and an initial SGD learning rate of 0.1 (decayed by a factor of 0.97 after each epoch). The training takes around three days on a server within 8 NVIDIA Tesla V100 GPU cards.


\subsection{Searched Architecture}
\begin{figure*}[h]
    \centering
    \includegraphics[width=\textwidth]{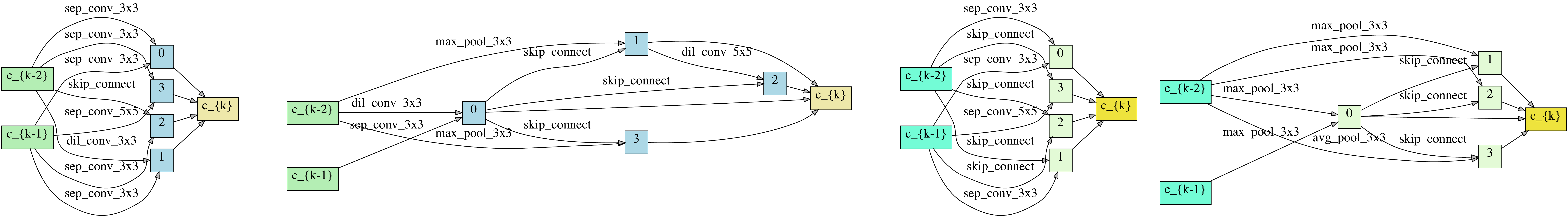}
    \caption{Searched Architectures. The left two sub-figures show an architecture that has an error rate of 2.34\% with a parameter size of 3.87M; The right two sub-figures show an architecture that has an error rate of 2.50\% with a parameter size of 2.86M.}
    \label{fig:cells}
\end{figure*}

Examples of the searched architectures are shown in Figure \ref{fig:cells}.

\section{Derivation of the MiLeNAS Second-Order Method}
In this section, we can derive a second-order method for MiLeNAS. As in DARTS, we also approximate $w^{*}$ by adapting $w$ using only a single training step:
\begin{equation*}
\nabla_{\alpha} \mmL_{\mathrm{val}}\left(w^{*}(\alpha), \alpha\right) \approx \nabla_{\alpha} \mmL_{\mathrm{val}}\left(w-\xi \nabla_{w} \mmL_{\mathrm{tr}}(w, \alpha), \alpha\right).
\end{equation*}
When applying the chain rule, we get
\begin{align*}
\nabla_{\alpha} \mmL_{\mathrm{val}}\left(w^{*}(\alpha), \alpha\right) \approx& \nabla_{\alpha} \mathcal{L}_{v a l}\left(w^{\prime}, \alpha\right)\\&-\xi \nabla_{\alpha, w}^{2} \mathcal{L}_{tr}(w, \alpha) \nabla_{w^{\prime}} \mathcal{L}_{val}\left(w^{\prime}, \alpha\right),    
\end{align*}
where $w^{\prime}=w-\xi \nabla_{w} \mathcal{L}_{t r a i n}(w, \alpha)$ denotes the weights for a one-step forward model.
Using the finite difference approximation, the complexity of the second order derivative in Equation 7 can be simplified. If we let $\epsilon^{val}$ be a small scalar 
and $w_{val}^{ \pm}=w \pm \epsilon^{val} \nabla_{w^{\prime}} \mathcal{L}_{v a l}\left(w^{\prime}, \alpha\right)$, then:
\begin{align*}
\nabla_{\alpha} \mmL_{\mathrm{val}}\left(w^{*}(\alpha), \alpha\right) \approx& \nabla_{\alpha} \mathcal{L}_{v a l}\left(w^{\prime}, \alpha\right) 
\\&- \xi \frac{\nabla_{\alpha} \mathcal{L}_{tr}\left(w_{val}^{+}, \alpha\right)-\nabla_{\alpha} \mathcal{L}_{tr}\left(w_{val}^{-}, \alpha\right)}{2 \epsilon^{val}}.    
\end{align*}

Following a similar derivation of $\nabla_{\alpha} \mmL_{\mathrm{val}}\left(w^{*}(\alpha), \alpha\right)$, we have
\begin{align*}
&\alpha =\alpha-\eta_{\alpha} \\ 
&\cdot\left[ \left( \nabla_{\alpha} \mathcal{L}_{val}\left(w^{\prime}, \alpha\right) - \xi \frac{\nabla_{\alpha} \mathcal{L}_{tr}\left(w_{val}^{+}, \alpha \right)-\nabla_{\alpha} \mathcal{L}_{tr}\left(w_{val}^{-}, \alpha \right)}{2 \epsilon^{val}} \right)  \right.\\ \nonumber
&\left. +\lambda \left( \nabla_{\alpha} \mathcal{L}_{tr}\left(w^{\prime}, \alpha\right) - \xi \frac{\nabla_{\alpha} \mathcal{L}_{tr}\left(w_{tr}^{+}, \alpha\right)-\nabla_{\alpha} \mathcal{L}_{tr}\left(w_{tr}^{-}, \alpha \right)}{2 \epsilon^{tr}} \right) \right],
\end{align*}
where $w^{\prime}=w-\xi \nabla_{w} \mathcal{L}_{\mathrm{tr}}(w, \alpha)$,  $w_{\mathrm{val}}^{ \pm}=w \pm \epsilon^{val} \nabla_{w^{\prime}} \mathcal{L}_{v a l}\left(w^{\prime}, \alpha\right)$, $w_{tr}^{ \pm}=w \pm \epsilon^{tr} \nabla_{w^{\prime}} \mathcal{L}_{tr}\left(w^{\prime}, \alpha\right)$.  $\epsilon^{tr}$ and $\epsilon^{val}$ are two scalars.

\section{Evaluation on \mldas-2nd} 
\label{sec:exp-milenas-2nd}

\begin{figure}[hbt!]
    \centering
    \begin{subfigure}{0.48\linewidth}
     \centering  
    \includegraphics[width=\linewidth]{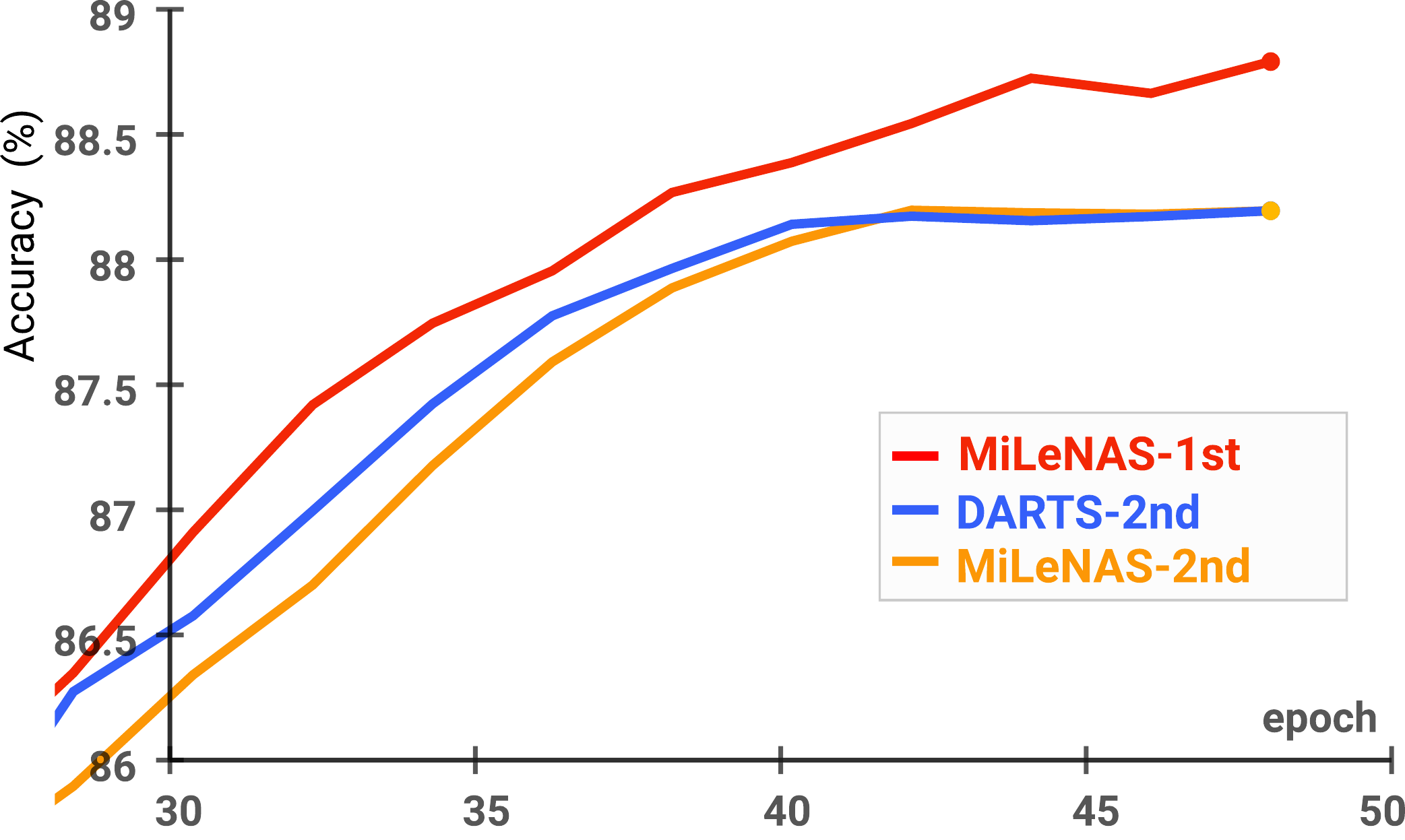}
    \caption{Validation Accuracy}
    \label{fig:validation_acc_v2_epoch}
    \end{subfigure}
    \begin{subfigure}{0.48\linewidth}
        \centering 
      \includegraphics[width=\linewidth]{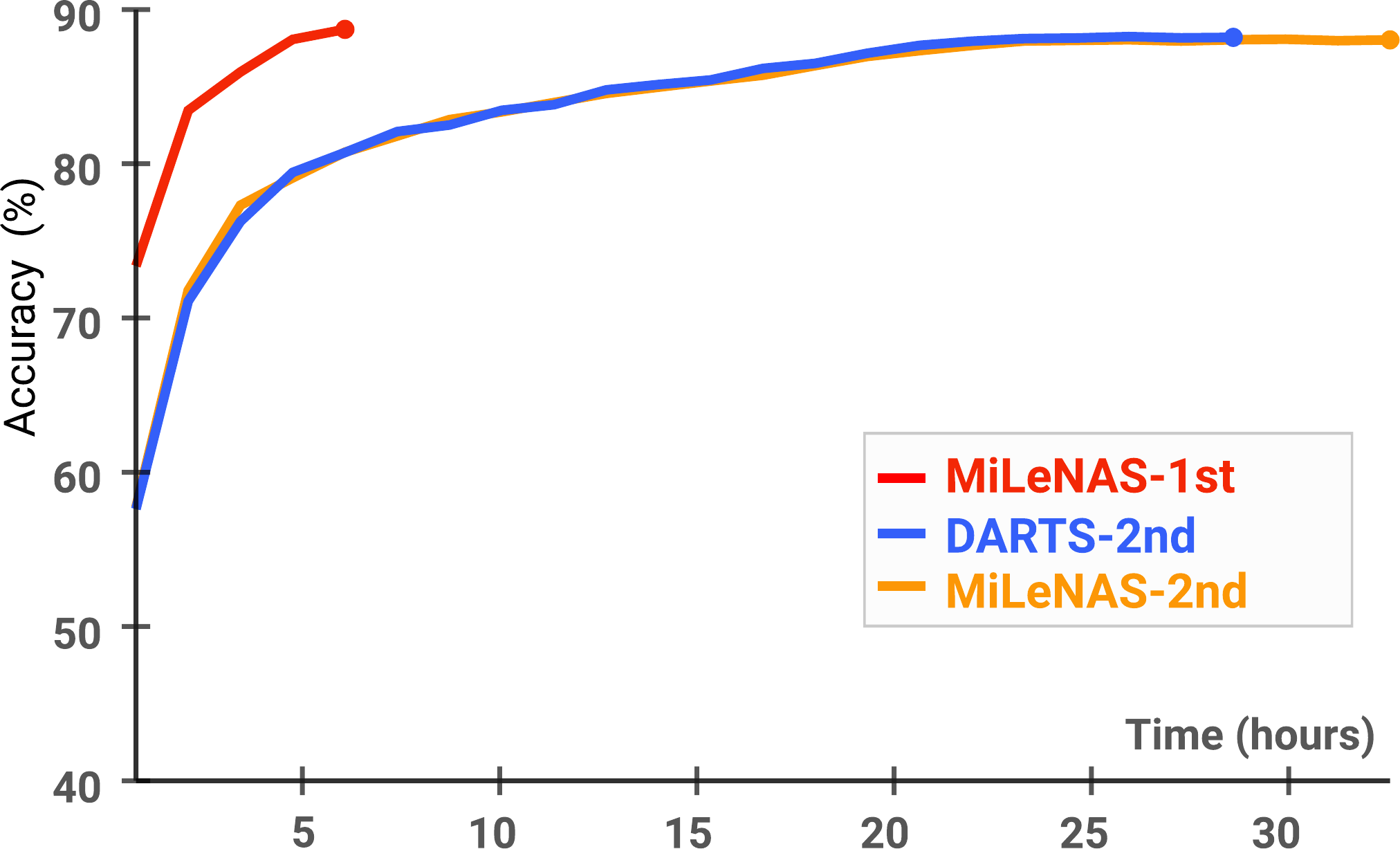}
    \caption{Searching Time}
    \label{fig:validation_acc_v2_epoch}
    \end{subfigure}
    \caption{Comparison between \mldas-2nd with \mldas-1st.}
    \label{fig:Searching_v2}
\end{figure}

As seen in our analysis, \mldas-2nd shows a similar gradient error and is also inefficient. To confirm this, we run experiments to compare its validation accuracy and training time with \mldas-1st. Each method is run four times, and the results are based on averages (Figure \ref{fig:Searching_v2}). The accuracy of \mldas-2nd is lower than that of \mldas-1st and similar to DARTS-2nd. The searching time of \mldas-2nd is the longest because it has one more inefficient term in the second-order approximation equation. Notably, in the early phase, \mldas-2nd is significantly less accurate than DARTS-2nd, which may be caused by the fact that \mldas-2nd has one more term with gradient error (refer to \mldas-2nd equation in Section 3.1). Thus,  among these methods, \mldas-1st is shown to be the optimal choice for addressing the NAS problem. 

\end{document}